\documentclass[a4paper,fleqn]{cas-sc}

\usepackage[square,sort&compress,numbers]{natbib}

\usepackage{lipsum}
\usepackage{siunitx}
\usepackage{subcaption}

\def\tsc#1{\csdef{#1}{\textsc{\lowercase{#1}}\xspace}}
\tsc{WGM}
\tsc{QE}

\begin{document}
\let\WriteBookmarks\relax
\def\floatpagepagefraction{1}
\def\textpagefraction{.001}

\shorttitle{End-user validation of BRIGHT with GUI for cervical cancer BT}

\shortauthors{L.R.M. Dickhoff et al.}  

\title [mode = title]{End-user validation of BRIGHT with custom-developed graphical user interface applied to cervical cancer brachytherapy}  



\author[1]{Leah R.M. Dickhoff}[orcid=0000-0001-6720-4380]
\cormark[1] 
\ead{L.R.M.Dickhoff@lumc.nl}

\author[1]{Ellen M. Kerkhof}[orcid=0000-0002-6070-7732]

\author[1]{Heloisa H. Deuzeman}[orcid=0000-0003-3282-8497]

\author[1]{Laura A. Velema}[orcid=0000-0003-3455-3407]

\author[1]{Stephanie M. {de Boer}}[orcid=0000-0002-8454-1597]

\author[1]{Lavinia A.L. Verhagen}[orcid=0009-0006-8738-4650]

\author[2]{Danique L.J. Barten}[orcid=0000-0002-0063-6899]

\author[2,3]{Bradley R. Pieters}[orcid=0000-0002-7427-8836]

\author[2,4]{Lukas J.A. Stalpers}[orcid=0000-0001-9574-0295]

\author[5]{Renzo J. Scholman}[orcid=0000-0003-2813-015X]

\author[5]{Pedro M. Matos}[orcid=0000-0002-3058-660X]

\author[5]{Anton Bouter}[orcid=0000-0003-4599-0733]

\author[1]{Carien L. Creutzberg}[orcid=0000-0002-7008-4321]

\author[5]{Peter A.N. Bosman}[orcid=0000-0002-4186-6666]

\author[1]{Tanja Alderliesten}[orcid=0000-0003-4261-7511]

\affiliation[1]{organization={Department of Radiation Oncology, Leiden University Medical Center},
    city={Leiden},
    country={The Netherlands}}

\affiliation[2]{organization={Department of Radiation Oncology, Amsterdam University Medical Centers (location University of Amsterdam)},
    city={Amsterdam},
    country={The Netherlands}}

\affiliation[3]{organization={Cancer Center Amsterdam, Imaging and Biomarkers},
    city={Amsterdam},
    country={The Netherlands}}

\affiliation[4]{organization={Cancer Center Amsterdam, Cancer Biology and Immunology},
    city={Amsterdam},
    country={The Netherlands}}

\affiliation[5]{organization={Evolutionary Intelligence Group, Centrum Wiskunde \& Informatica},
    city={Amsterdam},
    country={The Netherlands}}


\begin{abstract}
Multi-objective optimisation using BRIGHT has proven insightful and effective in prostate cancer brachytherapy treatment planning. BRachytherapy via artificially Intelligent GOMEA-Heuristic based Treatment planning (BRIGHT) generates multiple treatment plans, each with a different trade-off between tumour coverage and organs-at-risk sparing. BRIGHT was recently extended to cervical cancer brachytherapy. In this study, we present a novel, custom-developed graphical user interface (GUI) that enables plan navigation, pairwise comparisons, dose distribution visualisation, and possibility for adjustments — essential for efficient clinical use of BRIGHT.\par
End-user validation of BRIGHT with the dedicated GUI was conducted for cervical cancer brachytherapy by emulating clinical practice in ten previously treated patients. A multidisciplinary brachytherapy team used BRIGHT to create new treatment plans. GUI usability was assessed using the System Usability Scale (SUS). BRIGHT plan quality was compared to clinical practice via blinded one-on-one comparisons.\par
The GUI offered helpful features for plan navigation and evaluation, giving users quick insight into whether planning aims are achievable and what treatment options are available. The overall SUS score was 83.3, indicating an ‘excellent’ system. BRIGHT outperformed clinical practice in five out of ten patients regarding the coverage-sparing trade-off and performed equally well in the remaining five. The BRIGHT plan was preferred over the clinical plan in eight out of ten patients, four of which showed clinically relevant differences. The clinical plan was preferred in two patients, neither with clinically relevant differences.
In conclusion, BRIGHT, with its dedicated GUI, is a clinically viable and user-friendly tool for treatment planning in cervical cancer brachytherapy.
\end{abstract}

\begin{graphicalabstract}
\includegraphics[width=\linewidth]{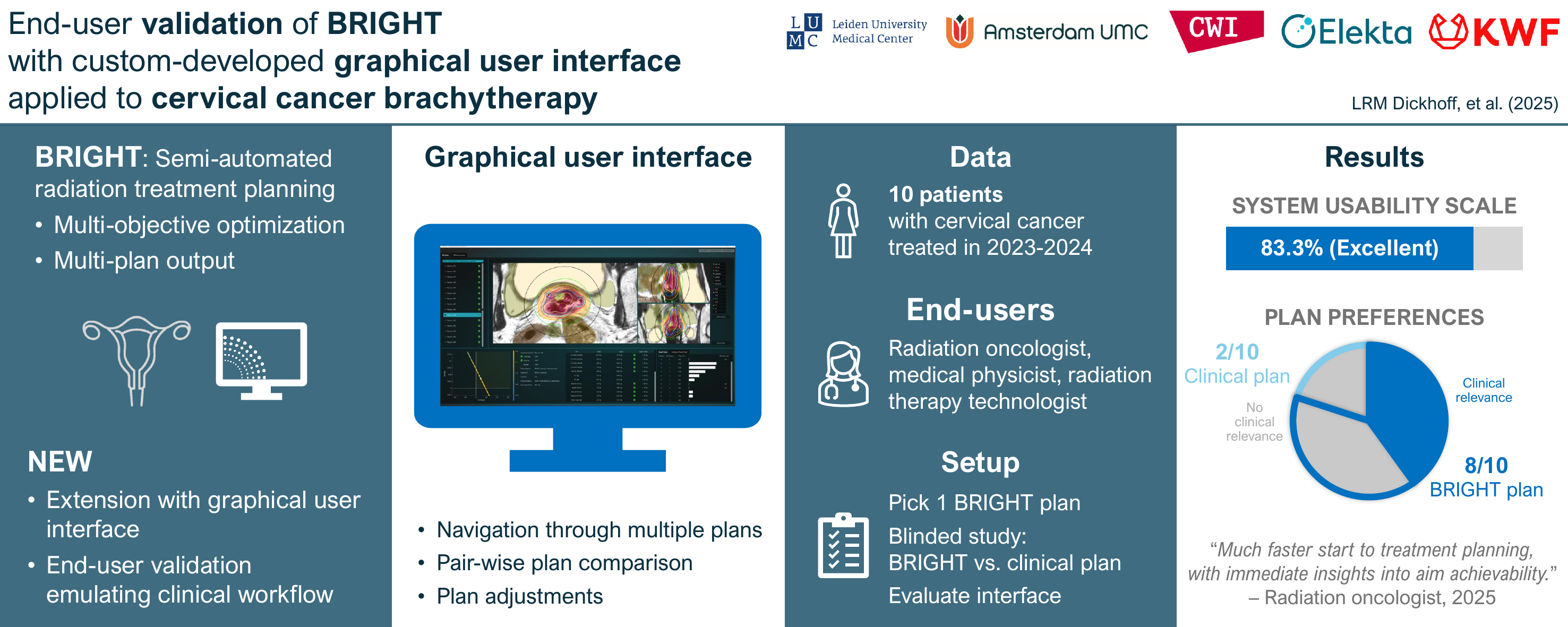}
\end{graphicalabstract}

\begin{highlights}
\item Semi-automatic brachytherapy treatment planning method called BRIGHT
\item First end-user validation of BRIGHT for cervical cancer brachytherapy
\item New graphical user interface developed for efficient use
\item BRIGHT plans preferred over clinical plans in 8/10 patients
\item BRIGHT interface aids plan navigation, immediate insight, and multiple plan options
\end{highlights}

\begin{keywords}
    Cervical cancer brachytherapy \sep
    End-user validation study \sep
    Graphical user interface \sep
    Semi-automated treatment planning \sep
    Multi-objective optimisation
\end{keywords}

\maketitle

\section{Introduction}

Brachytherapy (BT) is a mode of radiotherapy during which radioactive sources are placed in and around the tumour, and is commonly used as part of curative treatment for patients with cervical cancer \cite{Tanderup2014CurativeOptional}. In this article, we particularly consider high-dose-rate (HDR) BT, in which the placement of the sources is temporary and limited to the duration of the treatment. This is achieved through intracavitary and interstitial applicators, such as needles that are inserted into the body, and dedicated intracavitary applicators for cervical cancer BT.
A typical workflow of a BT treatment is shown in Supplementary Figure 1.
The radioactive source can reside at different dwell positions within these needles and applicators for a specific duration called dwell time. The goal is to deliver a curative minimum dose to the tumour cells, while at the same time limiting dose to the surrounding healthy organs at risk (OARs). Treatment planning thus consists of finding the best dwell positions, and the best dwell time for each of these dwell positions, in order to achieve both adequate dose coverage of the tumour and adequate sparing of OARs.

(Semi-)automated treatment planning in BT has been shown to outperform manual treatment planning both in terms of plan quality and planning time \cite{Maree2019EvaluationStudy, Shen2019IntelligentCancer, Rossi2024ClinicianBiCycle}. While methods generating one single treatment plan are preferred by some users due to reasons of simplicity and efficiency, other users prefer a multitude of plans to choose from in order to get insights into what is maximally achievable for the specific patient at hand \cite{Maree2019EvaluationStudy}. 
More specifically, the inherent multi-objective nature of tumour dose coverage versus sparing of OARs in radiotherapy can be captured by a multi-objective problem formulation. BRIGHT (BRachytherapy via artificially Intelligent GOMEA-Heuristic based Treatment planning) \cite{Barten2023TowardsBrachytherapy} is a semi-automated treatment planning method which distinguishes itself from other methods in that it directly optimises on dose-volume (DV) aims as given in a clinical protocol by grouping them in different objectives that are to be optimised simultaneously. These DV aims, together with the 3D dose distribution, represent the main plan evaluation metrics. In BRIGHT for prostate cancer, which is used in clinical practice, a two-objective problem formulation is used: one objective related to the coverage of the tumour, and one related to the sparing of OARs. A set of plans, called a Pareto approximation front, where each plan represents a different coverage-sparing trade-off is automatically generated in a few minutes. 
In BRIGHT for cervical cancer, these two objectives pertain to the coverage and sparing dose aims of the officially recommended EMBRACE-II protocol \cite{EMBRACE}. However, solely optimising on these does not lead to clinically acceptable treatment plans due to undesirable properties of the resulting 3D dose distribution \cite{Dickhoff2022AutomatedAims}. Therefore, a third objective was added to address such properties and to allow for the inclusion of potential clinic-specific preferences \cite{Dickhoff2022AdaptivePlanning}.

While the clinical acceptability of the automatically generated BRIGHT plans for cervical cancer BT has been stated during iterative feedback loops with local multidisciplinary BT treatment planning teams (in our clinic consisting of a radiation oncologist, a medical physicist, and a radiation therapy technologist), a direct comparison of the BRIGHT plans with the clinically used plans is essential to prelude a potential clinical introduction of BRIGHT for cervical cancer. In this work, we therefore present an end-user validation of BRIGHT in a retrospective setting, emulating our clinical practice when using BRIGHT to create treatment plans. It is the first to present such an end-user study, comparing (semi-)automated and clinical planning for cervical cancer BT, in which no clinical plan information was used for the optimisation of the automatically generated treatment plans. 
Other studies have demonstrated superiority of the (manually adjusted) automatically generated treatment plans, but
in those studies, the automated plans were explicitly calculated to match a key dose aim from the clinically used plans; specifically, the dose of radiation received by 90\% 
of the $\text{CTV}_{\text{HR}}$ (adaptive High Risk Clinical Target Volume of the primary tumour \cite{EMBRACE})
\cite{Rossi2024ClinicianBiCycle}. 
Clinically achieved dosimetric values are not available in clinical practice for new patients, potentially limiting the transferability of the published results.

For efficient use of BRIGHT in clinical practice, a graphical user interface (GUI) that enables navigation of the set of plans is essential. To the best of our knowledge, none of the currently commercially available treatment planning software systems support direct navigation through the Pareto approximation front resulting from multi-objective optimisation.
Some of the most commonly used external beam radiation treatment (EBRT) planning systems \cite{MCO-raystation, MCO-varian} use multi-criteria optimisation (MCO), which recognizes the multi-objective nature of the problem, but congregates the different aims into one objective using distinct weights. Thus, even though in MCO multiple criteria are accounted for, it is not a form of true multi-objective optimisation in that a single solution is found.
Fitting with this philosophy, the corresponding GUIs only show one plan at a time without providing insight regarding the relation of that plan to the other possible trade-off plans. Crucial is also that MCO is not (yet) available within these companies’ BT products. Moreover, some commercial software systems do not enable simple pairwise comparison between two plans \cite{oncentra-review}. 
Furthermore, BRIGHT for prostate cancer was in 2020-2024 used in the clinic by selecting three to five plans out of the set of plans \cite{Barten2021ArtificialExperience} without being able to look at the different plan-specific dose distributions before the selection is made. 
A visual check of the corresponding 3D dose distributions was performed in Oncentra Brachy (Elekta, Veenendaal, The Netherlands), which is not designed for this task, making it time-consuming.
While this approach proved effective — since BRIGHT improved the treatment plans enough to yield a gain in plan quality compared to the previously used manual planning — it did not leverage the full potential of BRIGHT because this requires a quick and intuitive comparison between plans before exporting the preferred plan for a final check.

We therefore developed a customized GUI that includes navigation through a Pareto approximation set of plans and pairwise comparisons, including visualization of dose distributions. It is integrated with the automated treatment planning method of BRIGHT
in order to automatically optimise a set of treatment plans for the patient at hand. The GUI development was finalized by conducting retrospective evaluation sessions with nine multidisciplinary BT professionals at our department in order to gather as much feedback as possible. 

In this work, for the first time, the custom-developed GUI is presented including features that are typically available in GUIs of BT treatment planning systems, such as dose distribution visualizations and a dwell time overview, as well as features unique to BRIGHT that enable direct interaction with the set of trade-off solutions. For ten patient cases, the GUI is evaluated according to usability metrics as part of an end-user validation of BRIGHT applied to cervical cancer BT during which a multidisciplinary BT team emulated clinical practice in a retrospective setting using BRIGHT with the custom-developed GUI to create a so-called BRIGHT plan. This study includes a blinded direct comparison of the created BRIGHT plans with the plans that were clinically used to treat the patients. It aims to evaluate the BRIGHT treatment plan quality and plan selection in the GUI for feasibility in a clinical setting.

\vspace{0.5\baselineskip}
\section{Methods}

\begin{figure}[h]
	\centering
		\includegraphics[width=\linewidth]{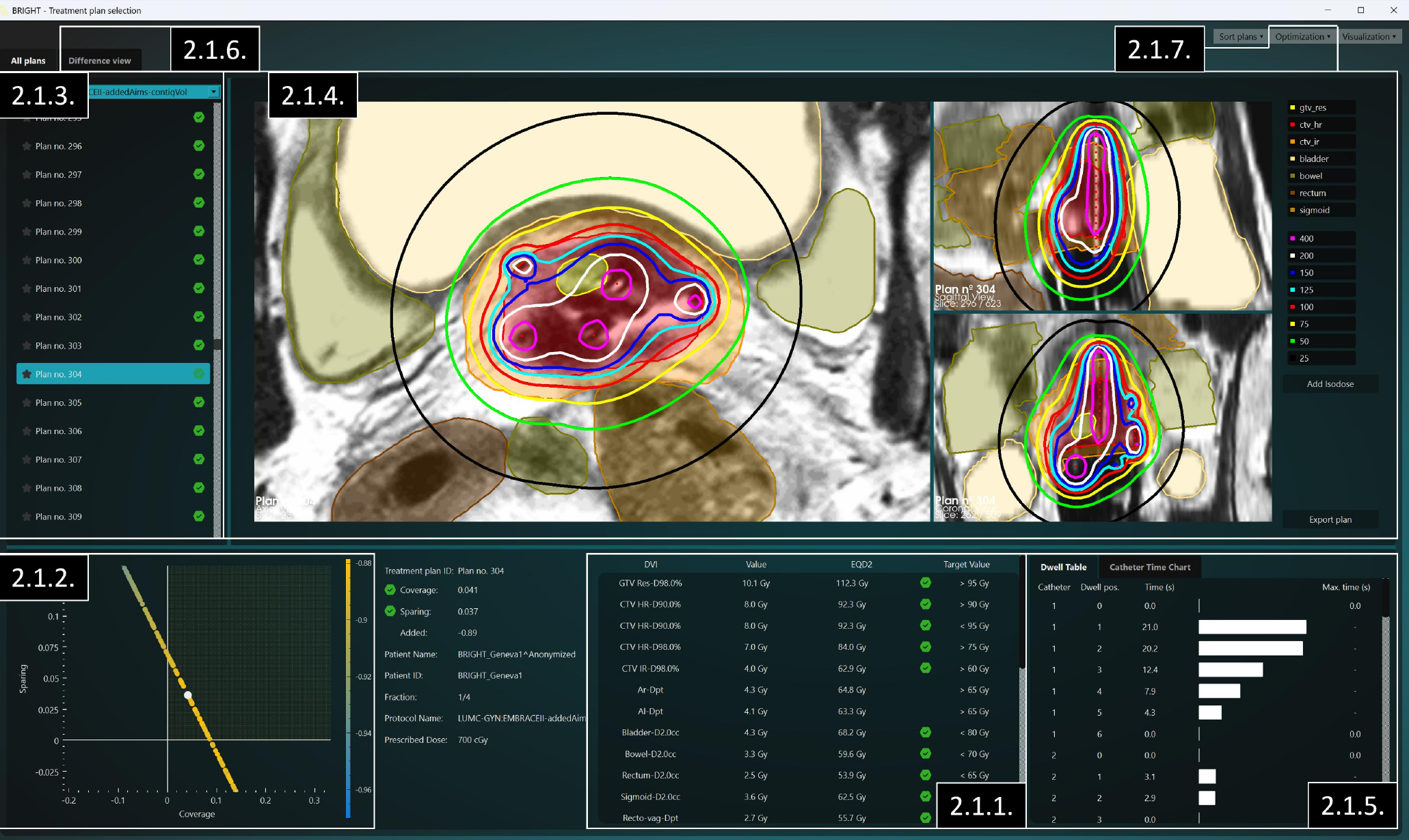}
	  \caption{Main window of the BRIGHT GUI. The numbered subparts refer to the subsections in which they are further explained. Subparts 2.1.2 and 2.1.3  are for treatment plan navigation and selection, whereas the other subparts aid in treatment plan evaluation. The three images show the different regions of interest as a delineated colorwash (legend: top right), superposed with isodose lines depicting areas receiving the same radiation dose (legend: mid right, in \% of the prescription dose of 7 Gy).}
        \label{fig:mainwindow}
\end{figure}

The methods section is comprised of two parts: the first part describes all available features in the GUI, of which the main window is shown in Figure \ref{fig:mainwindow}, whereas the second part addresses the metrics used for its evaluation as well as the study setup regarding the end-user validation and the comparison with current clinical practice.

\subsection{GUI Features} \label{sec:GUIfeatures}


\subsubsection{Dose volume metrics} \label{sec:DVtable}

Dose volume (DV) metrics, also referred to as DV indices (DVI), constitute the main quantitative manner of evaluating BT treatment plans. They relate to doses given to volumes. More specifically, a dose index $D_v$ refers either to the minimum dose associated with the most irradiated subvolume of $v$ $\si{\centi\metre\cubed}$, or to the minimum dose received by the $v\%$ most irradiated portion of the volume in the planned dose distribution. A volume index $V_d$ describes the subvolume which is planned to receive at least a dose of $d$ Gy, and $D_{\text{pt}}$ represents the dose to a specific point. DV metrics are defined for specific organs or regions of interest (ROIs), and for cervical cancer BT, DV aims are given in the EMBRACE-II protocol \cite{EMBRACE}.

BRIGHT directly uses dose calculation points - which are sampled uniformly at random in each of the ROIs - in order to compute and optimise the values of the DV metrics precisely without altering their mathematical definition. As opposed to this, most other automated treatment planning optimisation methods need to approximate the DV metrics as sigmoid functions to make them differentiable, since their optimisers rely on gradients.

In the GUI, DV values obtained for the currently selected treatment plan are given per ROI in a table. The planning aims in the EMBRACE-II protocol are given in total EQD2 (equivalent dose in 2 Gy fractions, in Gy), which corresponds to the summation of the given EBRT EQD2 dose, and the EQD2 of all given and planned $n$ BT fractions. These planning aims are given in the ’Target Value’ column in the GUI table, and the values associated with the selected treatment plan are presented in the ’EQD2’ column.  
The 'Value' column represents the physical dose $d$, equivalent to the amount of energy deposited by ionizing radiation per unit mass of tissue, without accounting for biological effects, measured in Gray (Gy) \cite{bt-dose-2021}, given for the single ($n=1$) planned fraction. It relates to the EQD2 per fraction by
\begin{equation*}
    \text{EQD2} = n d \frac{\alpha/\beta+d}{\alpha/\beta+2}
\end{equation*}
where $\alpha/\beta=3$ Gy for OARs and $\alpha/\beta=10$ Gy for cervix target volumes \cite{Potter2006RecommendationsRadiobiology, Thames1982ChangesRelationships}.
Whether the aim is achieved is indicated by a colored symbol: \includegraphics[height=7pt]{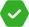} for an achieved aim, \includegraphics[height=7pt]{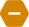} for an achieved limit (less strict than the aim, as given in the EMBRACE-II protocol \cite{EMBRACE}), and \includegraphics[height=7pt]{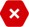} when neither aim nor limit are reached. There is no symbol next to the A points, as they were not included in the optimisation as doing so resulted in unnatural dose distributions. Therefore, the doses at these points are only reported for evaluation purposes.

For semi-automated treatment planning, optimising on solely the DV metrics with associated planning aims from the EMBRACE-II protocol was not sufficient to lead to clinically acceptable treatment plans \cite{Dickhoff2022AutomatedAims}. Therefore, in an iterative feedback loop with local multidisciplinary BT teams, additional DV metrics with associated planning aims were tuned in order to tackle unwanted properties in the 3D dose distributions and thereby obtain clinically acceptable plans \cite{Dickhoff2022AdaptivePlanning}. Reference points, as well as these added DV metrics, are listed in the table following the DV metrics from the EMBRACE-II protocol, as shown in the extended version in Supplementary Figure 4.

\subsubsection{Pareto approximation front} \label{sec:paretoapprox}

The DV metrics for cervical cancer BT can be subdivided into three different categories:
\begin{itemize}
    \item Coverage aims: DV aims from the EMBRACE-II protocol, relating to target volume coverage, that require maximization,
    \item Sparing aims: DV aims from the EMBRACE-II protocol, relating mostly to OAR sparing, that require minimization,
    \item Added aims: Additionally tuned aims which were deemed necessary to obtain clinically acceptable treatment plans \cite{Dickhoff2022AdaptivePlanning}, and can potentially include institution-specific preferences \cite{Dickhoff2025MedPhys}.
\end{itemize}

A suitable choice for optimising treatment plans is therefore multi-objective optimisation, where each objective is associated with one of the above categories. The multi-objective real-valued gene-pool optimal mixing evolutionary algorithm (MO-RV-GOMEA) \cite{mo-rv-gomea-main} is used in BRIGHT, as it has been proven efficient for such optimisation \cite{Luong2018ApplicationTreatment, Maree2019EvaluationStudy, Milickovic2001ApplicationAlgorithms}. In BRIGHT, this population-based method optimises a whole set of treatment plans at once, where each plan describes a different best possible trade-off between the objectives. Pareto dominance is used to compare solutions in a multi-objective fashion, where one solution dominates another if it is better in one objective, and at least as good in all other objectives. 
Each objective in BRIGHT, being the Least Coverage Index (LCI), Least Sparing Index (LSI), and Least Added Index (LAI), is a sum of the difference between the current value $\text{DV}^a$ and its aim $\text{DV}^a_\text{aim}$ over all respective aims $a$, which is computed for each plan $p$: 
\begin{flalign}
\begin{split} \label{eq:lcilsi}
    \mathrm{LCI}_w(p) &= \sum_{a \ \in \ \text{coverage aims}} w_a(\text{DV}^a-\text{DV}^a_\text{aim}), \\
    \mathrm{LSI}_w(p) &= \sum_{a \ \in \ \text{sparing aims}} w_a(\text{DV}^a_\text{aim}-\text{DV}^a), \\
    \mathrm{LAI}_w(p) &= \sum_{a \ \in \ \text{added aims}}
    \begin{cases}
    w_a(\text{DV}^a-\text{DV}^a_\text{aim}) & \text{if $a$ is a coverage aim},\\
    w_a(\text{DV}^a_\text{aim}-\text{DV}^a) & \text{if $a$ is a sparing aim}.
    \end{cases}
\end{split}
\end{flalign}
It actively optimises the DV metric of which the value is currently furthest away from its aim (i.e., the worst-case) in each objective, by dynamically adjusting the weights $w_a$ throughout the optimisation  \cite{Bouter2019GPU-acceleratedBrachytherapy}. The non-weighted LCI and LSI correspond to the difference in the worst-case DV metric: $\mathrm{LCI}(p)=\min_{a \in \text{coverage aims}} \left( \text{DV}^a-\text{DV}^a_\text{aim} \right)$ and $\mathrm{LSI}(p)=\min_{a \in \text{sparing aims}} \left( \text{DV}^a_\text{aim}-\text{DV}^a \right)$.

When all coverage aims are satisfied, then $\mathrm{LCI}(p)\geq0$, and when all sparing aims are satisfied, then $\mathrm{LSI}(p)\geq0$. Thus, plans in which all aims from the EMBRACE-II protocol are satisfied ($\mathrm{LCI}(p)\geq0$ and $\mathrm{LSI}(p)\geq0$) are immediately recognizable by being in the so-called Golden Corner. In the BRIGHT GUI, it has therefore been chosen to represent the coverage and sparing objectives in a 2D graph, with the first quadrant marked as the Golden Corner. This decision is supported by the fact that, as the objectives are currently tuned, there is a strong correlation between the $\mathrm{LSI}_w$ and the $\mathrm{LAI}_w$, resulting in a line instead of a sphere octant of treatment plans. The value of the third objective is represented on a yellow (higher values) to blue (lower values) colour scale. Since achievable values were found to differ vastly between different patient cases, 
and relative values of the added aims within the LAI are more important than the absolute values,
the exact value of the $\mathrm{LAI}_w$ is not of importance but may merely serve as a comparison between different plans. The displayed set of treatment plans (i.e., the Pareto approximation front) ranges from high sparing (top left) to high coverage (bottom right), which allows for intuitive navigation to select the preferable plan for the patient at hand.

Every dot in the graph corresponds to one treatment plan, and the user can select a treatment plan to be shown by clicking on a dot. The currently selected plan is white. By default, when the GUI is opened, the plan with the most equally balanced trade-off value according to $ p = \operatorname*{argmax}_{p \in plans}\left[\min\left(\text{LCI}(p), \text{LSI}(p)\right)\right]$ is shown.

\subsubsection{Plan list} \label{sec:planlist}

The set of plans is also given in the plan list. When multiple optimisation runs have been done (see, e.g., Section \ref{sec:reopt} below), then there are multiple sets of plans to choose from, and the currently shown set of plans is selectable in the drop-down menu above the list of plans. This allows the user to switch between different sets.

The plan list is by default sorted from high sparing to high coverage, in the same order as the plans are in the Pareto approximation front visualized in \ref{sec:paretoapprox}. There is also an option to sort the plans based on how much the needles contribute to the total dwell time (in all catheters, i.e., needles and applicator), by selecting 'Sort plans' and then 'By needle contribution' in the menu on the top right.

The plan list allows the user to efficiently switch between different plans while reviewing specific image slices of interest with dose distributions overlaid, and thereby select the plan that best represents a certain desirable dose distribution property. Finally, plans can be marked as favorite by enabling the star located at the left of the plan number in the list. This plan will then also become visible as a yellow star in the Pareto approximation front. Favorited plans can be compared one-on-one in a separate tab (see Section \ref{sec:diffview}).

\subsubsection{Dose distributions} \label{sec:dosedistr}

The different delineated ROIs (see Supplementary Figure 1 visualizing the BT workflow steps including delineation) are visualized using a colourwash on the shown image slice. The dose distribution of the selected plan is displayed on top of this in three views: the axial, sagittal, and coronal views. 
The three views can be displayed along the image, applicator, or needle axes.
Double-clicking on either view will maximize that view. Magnetic resonance imaging (MRI), computed tomography, and ultrasonography images can be displayed.

Dose values are calculated according to the AAPM TG-43 dose calculation formalism \cite{Rivard2004UpdateCalculations}. Specific isodose lines, i.e., lines which show areas receiving the same radiation dose in the plan, can be added with the 'Add Isodose' button under the legend. Each of the displayed ROIs, as well as the different isodose lines can be toggled on/off by clicking on the respective element in the legend.

The different dwell positions are shown by red circles, hovering over a specific position will display its catheter number as well as its dwell position number within that catheter. Furthermore, a plan can be exported by selecting the 'Export plan' button under the legend (which outputs an RTPlan DICOM file \cite{dicom-obj-def}). 

\subsubsection{Dwell time table} \label{sec:dwelltable}

Two tabs are selectable in the dwell time table section. The first tab displays the dwell times for each of the dwell positions in the selected plan (see Figure \ref{fig:mainwindow}). The different columns correspond to the catheter number, dwell position number within that catheter, time (in s), and visualization as a bar. These are available in most commercially available software systems, since they represent an important overview of all dwell times of the whole treatment plan.
By default, the minimum and maximum allowed time for any used dwell position is set to 1 s and 150 s, respectively.
In the BRIGHT GUI dwell time table, a column has now been added that represents the maximum allowed dwell time (in s) per dwell position. The feature allows for limiting the dwell times for specific positions, or positions can even be disabled by setting the maximum to 0 s. After altering the desirable maximum time(s), the user should select the 'Apply' button, which re-optimises the treatment plans (see Section \ref{sec:reopt}), and gives a new set of treatment plans, with new optimised trade-offs between the objectives, optimised based on the newly restricted dwell times.

\begin{figure}[h]
	\centering
		\includegraphics[width=0.4\linewidth]{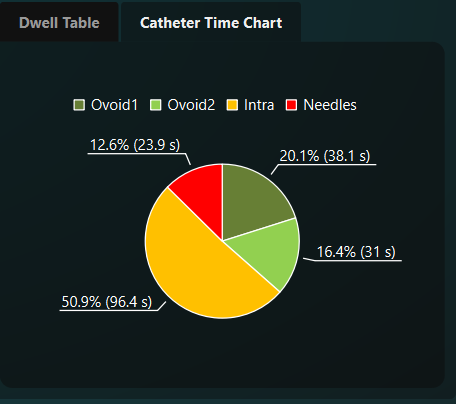}
	  \caption{Catheter time chart with distribution of dwell times associated with the separate ovoids, the intrauterine applicator, and the needles.}
        \label{fig:taart}
\end{figure}

The second tab is called 'Catheter time chart' (see Figure \ref{fig:taart}) and allows for a quick overview of the total dwell time distribution in the currently selected plan, by giving the percentages and total time (in s) that stem from the separate ovoids, the intrauterine applicator, and the needles (see Supplementary Figure 2 for a visualization of the different parts).

\subsubsection{Difference view} \label{sec:diffview}
\begin{figure}[h]
	\centering
		\includegraphics[width=\linewidth]{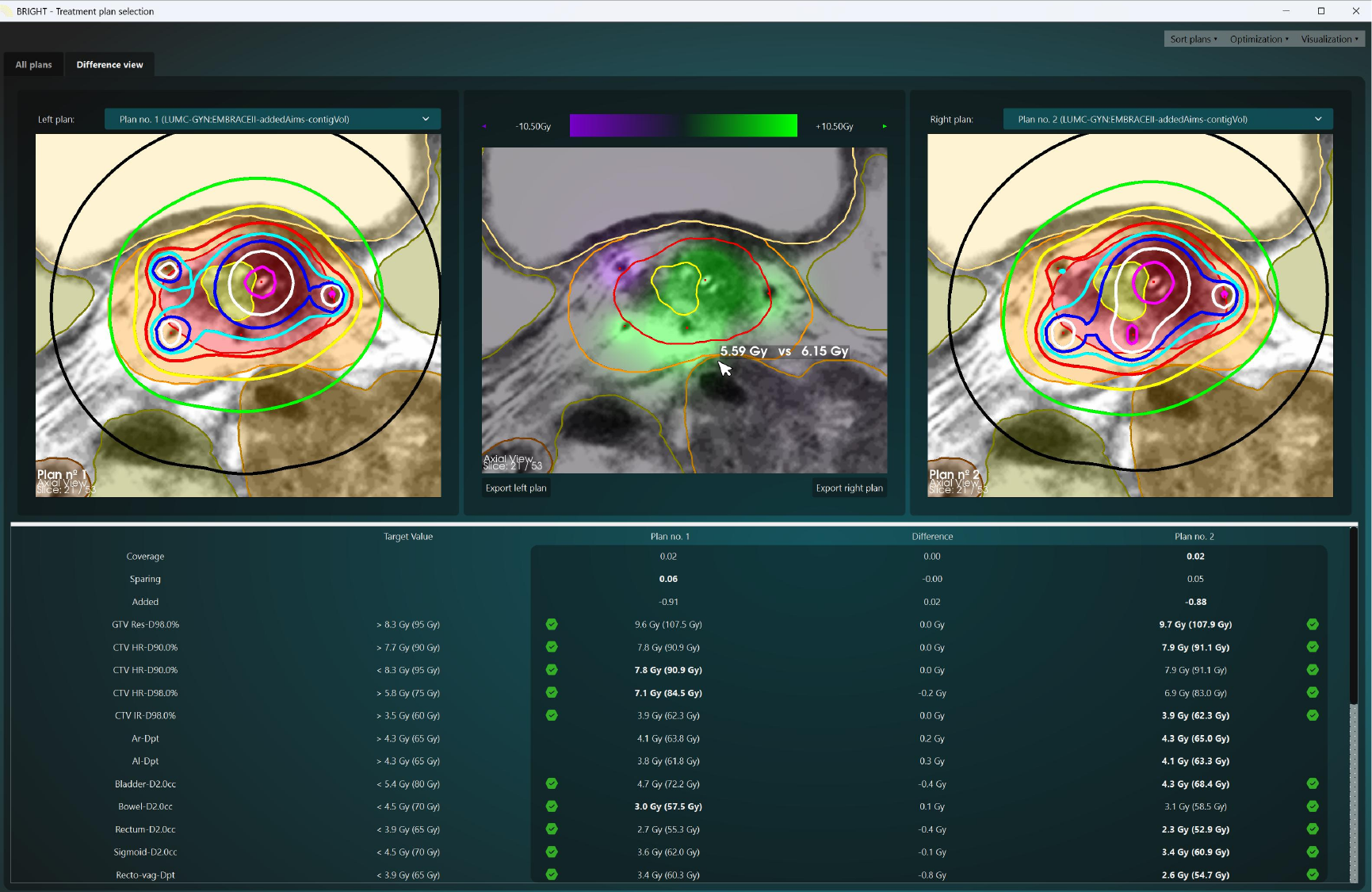}
	  \caption{
      Tab in the BRIGHT GUI allowing for one-on-one plan comparison with a (retractable) middle view corresponding to the difference in dose between the two plans as a colourwash, with green representing a higher dose from the right plan, and purple a higher dose from the left plan. All dose aims from EMBRACE-II are given side by side in a table, in physical dose, and total EQD2 in brackets.
      The legend of the ROIs and isodose lines can be found in Figure \ref{fig:mainwindow}}
        \label{fig:diffview}
\end{figure}

Plans which have been marked as favourite in the plan list can be compared to each other in the difference view, which is a separate tab selectable in the top left corner. It is displayed in Figure \ref{fig:diffview} and shows the dose distributions of the two selected plans on each side, with a colourwashed difference view in the middle that indicates exactly where the right (in green), respectively left (in purple), plan contains a higher dose area relative to the other plan. Hovering with the mouse cursor allows for an even more precise comparison since it results in displaying the dose (in Gy) that the right and left plan obtain in the current voxel. This middle column can be hidden (with the ‘Visualization’ button), should the user wish to see the two plans directly side by side.

Below the dose distributions, there is a table with the different DV metrics and obtained values associated with both plans (similarly to Section \ref{sec:DVtable}), along with the target value (in physical dose and in total EQD2 in brackets) as a reference. The bold values are the better values for that DV metric. In a column between the values pertaining to each plan, their difference in physical dose is given.

\subsubsection{Re-optimisation} \label{sec:reopt}
\begin{figure}[h]
	\centering
		\includegraphics[width=0.55\linewidth]{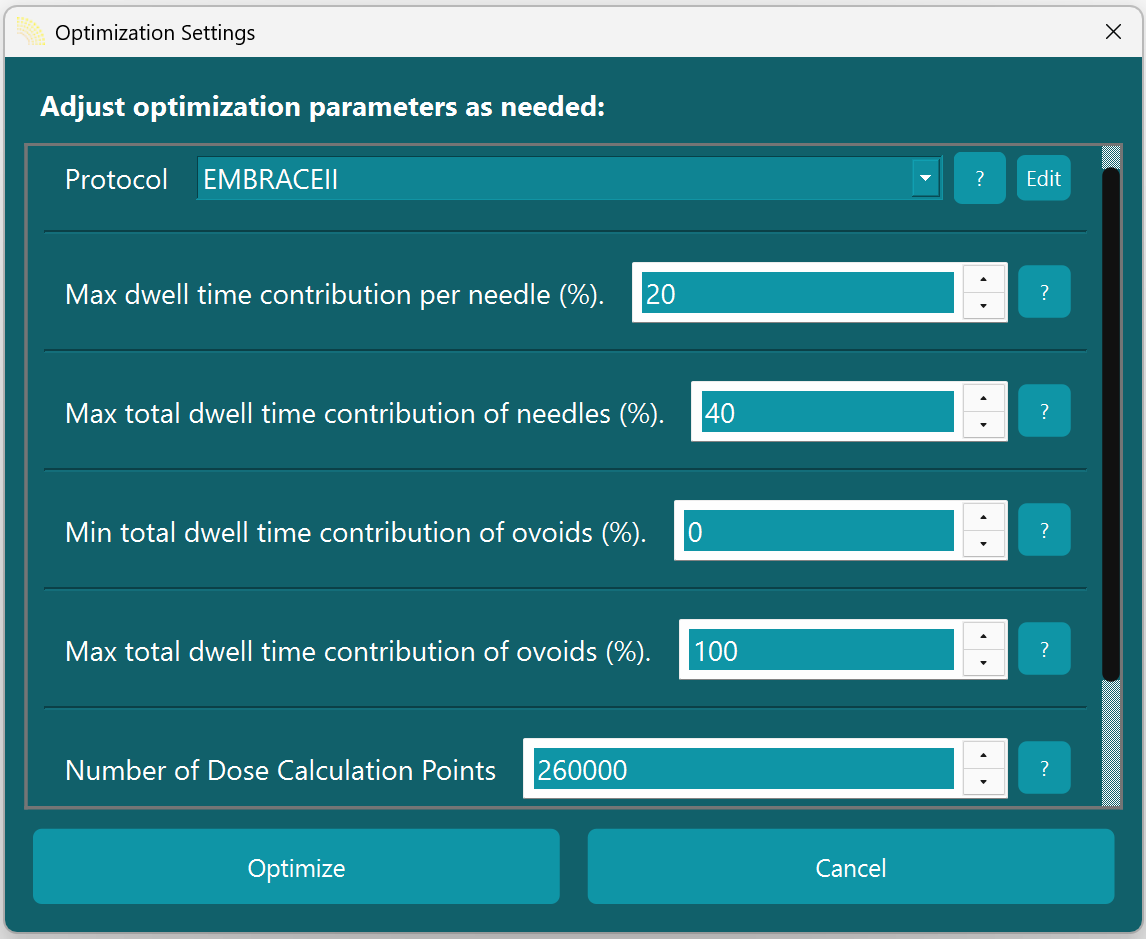}
	  \caption{Re-optimisation settings window.}
        \label{fig:reopt}
\end{figure}

Should the set of plans need some further tuning, e.g., a specific OAR needs to be spared more, or the contribution from the needles to the total dwell time (of which the default is 20\% for one needle and 40\% for all needles) is desired to be adjusted, then the feature to re-optimise the set of plans can be used. Selecting the button 'optimisation' on the top right opens a window as shown in Figure \ref{fig:reopt}. The maximum 
(and minimum) 
contribution of one single needle, of all needles together, and of the ovoids 
can be adjusted here. Moreover, the individual DV aims can be adjusted by clicking on the 'Edit' button next to the protocol name. Re-optimisation does not start from scratch, but uses the already optimised treatment plans in order to continue optimisation, allowing for a new set of treatment plans to be optimised in just 30 s \cite{Dickhoff2024KeepingBrachytherapy, Bakker2022Warm-startingTime}. The initial set of treatment plans is also kept in order to allow the user to compare plans between the distinct sets.

\vspace{\baselineskip}
\subsection{Validation study} \label{sec:eval}

The end-user validation study consists of two phases: the validation phase during which the GUI 
was evaluated, and the comparison of BRIGHT with clinical practice.

\subsubsection{Data} \label{sec:data}

The plan evaluation study includes coded planning and treatment data from 10 patients with cervical cancer, treated with
four fractions (in two BT procedures) with 7 Gy specified at the 100\% isodose surrounding the $\text{CTV}_\text{HR}$, aiming at a $\text{CTV}_\text{HR}$ EQD2 $\text{D}_{90\%}$ of 90-95 Gy.
Their BT treatment was preceded by EBRT with a dose schedule of 
45 Gy in 25 fractions of 1.8 Gy using a Volumetric Modulated Arc Therapy technique with library of plans \cite{vmat-libplans}.
In order not to include outliers in the small patient set, 
patient cases were judged to be representative by two radiation therapy technologists who reviewed the corresponding MRI scans which were made after implantation of the applicator and needles. 
The patient set comprised ten consecutively treated patients treated in 2023-2024 with the Geneva\textsuperscript{TM} universal gynaecological applicator (Elekta, Stockholm, Sweden; see Supplementary Figure 2), with the inclusion criteria listed above.
Only the first fraction per patient case was considered in order to include as many different cases as possible.
Included cases had 
T stages according to the 2021 Tumor-Node-Metastasis (TNM) staging system of T1b3 (1), T2 (1), T2b (7), T3b (1), and four patients had pelvic lymph node involvement (N+).
They had a median number of needles of 5 (range: 4-7), a mean $\text{CTV}_\text{HR}$ volume of 34.5 \si{\centi\metre\cubed} (range: 15.1-87.3), and a mean $\text{GTV}_\text{res}$ (Residual Gross Tumour Volume of the primary tumour \cite{EMBRACE}) volume of 8.9 \si{\centi\metre\cubed} (range: 0.8-31.8).

The plans utilized in clinical practice underwent manual optimisation using Oncentra (Oncentra Brachy, version 4.6.0, Elekta, Veenendaal, The Netherlands). This involved starting from a library plan \cite{gecestro-tps}, where only applicator dwell positions are used and normalised 
to the A points.
Subsequently, the plan underwent alterations by adjusting dwell times of selected dwell positions and/or manual optimisation was performed by dragging and dropping isodose lines. The final values obtained for the DV metrics are checked by copy-pasting the physical dose per fraction as given in Oncentra to an Excel spreadsheet, in which the total EQD2 is summed up over all four fractions (added to the EBRT EQD2 dose), and compared to the EMBRACE-II planning aims. This treatment planning process is preluded by manual delineations of the ROIs in RayStation (version 10B) by RaySearch Laboratories. The delineations are then transferred to Oncentra, where the applicator and needles have been reconstructed (see Supplementary Figure 1).

\subsubsection{End-users}
A multidisciplinary BT treatment planning team consisting of a radiation oncologist, a medical physicist, and a radiation therapy technologist performed treatment planning by use of BRIGHT in a retrospective setting. A multidisciplinary team was selected to closely emulate our clinical practice, and to assess BRIGHT and the GUI from the perspective of each discipline. 

\subsubsection{Optimisation settings}
BRIGHT was run for 3.7 min (on an NVIDIA GeForce RTX 4080 Mobile GPU), proven necessary for approximate mathematical convergence \cite{Dickhoff2025MedPhys}, though speed-ups are still possible (see Section \ref{sec:discussion}). A total of 260,000 dose calculation points (20,000 per ROI) are used during optimisation, while the final results that are visualized in the GUI are re-evaluated on 500,000 points \cite{Dickhoff2022AdaptivePlanning}. A contiguous volume constraint was used to ensure a single 250\% isodose volume (that is larger than 0.125 \si{\cubic\centi\metre}), ensuring a more homogeneous dose distribution that avoids multiple large high-dose regions. Worth noting is that optimising with contiguous volumes requires the use of a graph-based discretization, which is a capability distinct to BRIGHT because it directly uses discretizable dose calculation points, whereas other methods rely on differentiable and continuous gradients. 

Following clinical practice, some dwell positions are deactivated for optimization initially. These are the most distal and most proximal dwell positions in both ovoids, the most caudal dwell position in the intrauterine applicator, all except two dwell positions outside of the $\text{CTV}_{\text{IR}}$ (Intermediate Risk Clinical Target Volume of the primary tumour \cite{EMBRACE}) cranially in the intrauterine applicator, and all needle dwell positions that are further than 4.39 \si{\milli\metre} away from the $\mathrm{CTV_{HR}}$ \cite{Dickhoff2025MedPhys}. BRIGHT can use all other dwell positions freely, without prior (de)activation as it was done for the clinical plan.

\subsubsection{Study setup} \label{sec:studysetup}

In order to prepare the end-users for this study, each end-user participated in a training session before the start of the validation sessions, in which all features of the BRIGHT GUI were covered and a hands-on, step-by-step exercise was included to ensure they were comfortable using the software.

\begin{figure}[h]
	\centering
		\includegraphics[width=0.8\linewidth]{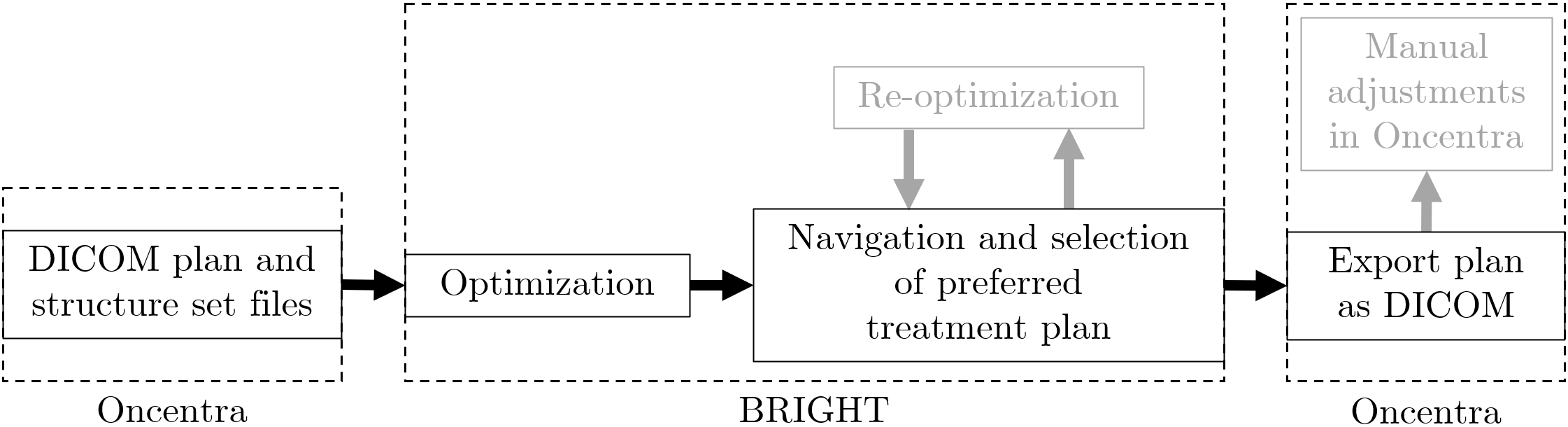}
	  \caption{Workflow for BRIGHT treatment plan selection (and adjustments). Steps shown in grey are optional. See \cite{dicom-obj-def} for explanations of the DICOM standard.}
        \label{fig:workflow}
\end{figure}

Each validation session was dedicated to the treatment planning process of one patient case, with a maximum duration of 1h was scheduled. The actual time required was recorded and is reported in the results. The different steps are illustrated in Figure \ref{fig:workflow}. First, the BT team was presented with clinical patient information since this information could have potential impact on the personalized treatment plan selection in clinical practice. This information contains the clinical drawing (which depicts the extent of disease based on the gynaecological examination at diagnosis) and the first consult notes, such as patient age, tumour stage and location, possible parametrium invasion, full undergone treatment for cervical cancer, and potential toxicities. Second, BRIGHT optimisation was performed. Third, the BT team used the GUI, utilizing the features they deem necessary, in order to finally select one preferred treatment plan in BRIGHT. Fourth, the chosen treatment plan was exported from BRIGHT and imported into Oncentra. This serves as a final check of the plan in already regulatory-approved software, and to allow for potential manual adjustments - should the BT team wish to adjust the BRIGHT-optimised treatment plan. Finally, after each patient, the BT team was asked to, together, fill out Questionnaire 1 (see Supplementary Table 3) consisting of different questions on treatment plan quality and GUI features used. The whole session was registered using screen and audio recording.

\subsubsection{GUI usability} \label{sec:GUIusability}
After all BRIGHT treatment plans were finalized, the end-users (one radiation oncologist, one medical physicist, and one radiation therapy technologist) were asked to separately fill in Questionnaire 2 with three different types of questions. The Likert scale questions are presented in Table \ref{tab:SUSquestionnaire}.
Firstly, the end-users' perceived usability was assessed by using the System Usability Scale (SUS) \cite{Brooke2020SUS:Scale} (Questionnaire 2A). It consists of 10 questions with five answer options ranging from 1 = strongly disagree to 5 = strongly agree. A SUS score of 68 is considered above average, whereas the perfect score is equivalent to 100. In order to make the numeric score more explicit, an adjective mapping was used, indicating that a score of above 71.1 is considered good, and a score above 80.8 is excellent \cite{Bangor2009DeterminingScale, Sauro20185Score}.

Secondly, four more software-specific questions were included  (Questionnaire 2B), designed in the same format as the SUS questions, using the same five answer options. Thirdly, four open-ended questions (Questionnaire 2C, see Supplementary Table 4) were designed to evaluate whether the end-users perceived added value in using BRIGHT compared to commercial BT treatment planning software. These questions also addressed which features were perceived as satisfiable, missing, or in need of improvement. Responses to the open questions were evaluated using reflexive thematic analysis 
 \cite{Braun2006UsingPsychology, Braun2019ReflectingAnalysis}.

\subsubsection{Comparison with clinical practice}

For a direct one-on-one comparison between the plan created by use of BRIGHT and approved by the team (after the steps as laid out in Section \ref{sec:studysetup}), which we will refer to as the BRIGHT plan, and the clinically used plan for each patient case (as described in Section \ref{sec:data}), four radiation oncologists separately compared the plans. The decision was made to only include radiation oncologists since they also have the responsibility for the treatment plan in clinical practice. A five-week interval was left in between the creation of the last BRIGHT plan and the first session of the comparison study to avoid biasing. Patient information which was available during creation of the BRIGHT plan and the clinical plan was not provided again prior to the comparison. The study was blinded; therefore, the BRIGHT and clinical plans were randomly assigned to plan numbers 1 and 2.

Using the BRIGHT difference view tool (see Section \ref{sec:diffview}), the radiation oncologists were asked to indicate which plan they prefer, expressing a strong, weak, or no preference, based on, first, the DV metric values; second, a visual inspection of the dose distribution while scrolling through the slices; and finally, their overall choice of the treatment plan they would use to treat the patient. The radiation oncologists were also asked 
whether the perceived differences could potentially be clinically relevant in terms of patient outcome, including tumour control, toxicity, or both.
The corresponding questionnaire (Questionnaire 3) can be found in Supplementary Table 5. An overview of all questionnaires used in this work can be found in Table \ref{tab:questionnaires}. In case of interobserver variation, the relevant patient cases were discussed during a consensus meeting in which all four radiation oncologists were asked to agree a single overall plan choice. This meeting took place three weeks after the final individual comparison session.

Finally, following each one-on-one comparison, the four end-users were asked, independently, whether they think either plan was clinically acceptable, with a yes or no answer. This was also done blinded and the question was asked separately for each of the two plans.

\begin{table}[h]
\footnotesize
    \centering
    \begin{tabular}{m{0.5cm}m{5.5cm}m{1.9cm}m{3.5cm}m{2.5cm}}
        No. & Purpose & End-user(s) & Filled in at time & Full list of questions \\\hline\hline
        1 & Patient-specific: BRIGHT plan quality, features used & BT team & After each patient case validation session & Suppl. Table 3 \\\hline
        2A & SUS & \multirow{3}{*}{\begin{tabular}{@{}l@{}}Every end-user\\separately\end{tabular}} & \multirow{3}{*}{After all validation sessions} & Table \ref{tab:SUSquestionnaire} \\
        2B & Software-specific & & & Table \ref{tab:SUSquestionnaire} \\
        2C & Open questions & & & Suppl. Table 4 \\\hline
        3 & Comparison of BRIGHT plan vs. clinical plan & Rad. oncologist & During comparison study & Suppl. Table 5
    \end{tabular}
    \caption{Overview of all questionnaires used in this work.}
    \label{tab:questionnaires}
\end{table}

\subsubsection{Preceding final tuning phase}

Prior to performing the end-user validation, as part of the developmental iterative feedback loop, a final tuning phase of the new version of BRIGHT was done with all BT professionals working in our department. The setup of the associated sessions was similar to the end-user validation sessions described in Section \ref{sec:eval}, and resulted in the GUI as described in Section \ref{sec:GUIfeatures}. 
Changes that were made based on the received feedback are presented in Supplementary Material B.2.1.

\vspace{0.5\baselineskip}
\section{Results}

\subsection{Emulated treatment planning with BRIGHT} \label{sec:results:planning}

\begin{figure}[h]
    \centering
    \begin{minipage}{0.71\textwidth}
        \raggedleft
        \includegraphics[width=0.95\textwidth]{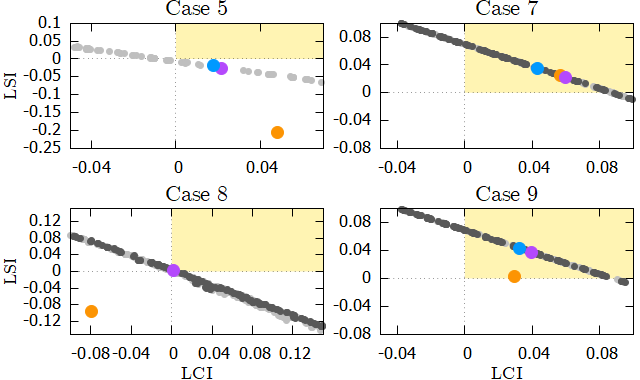}
    \end{minipage}
    \hspace{6pt}
    \begin{minipage}{0.26\textwidth}
        \raggedright
        \includegraphics[width=0.9\textwidth]{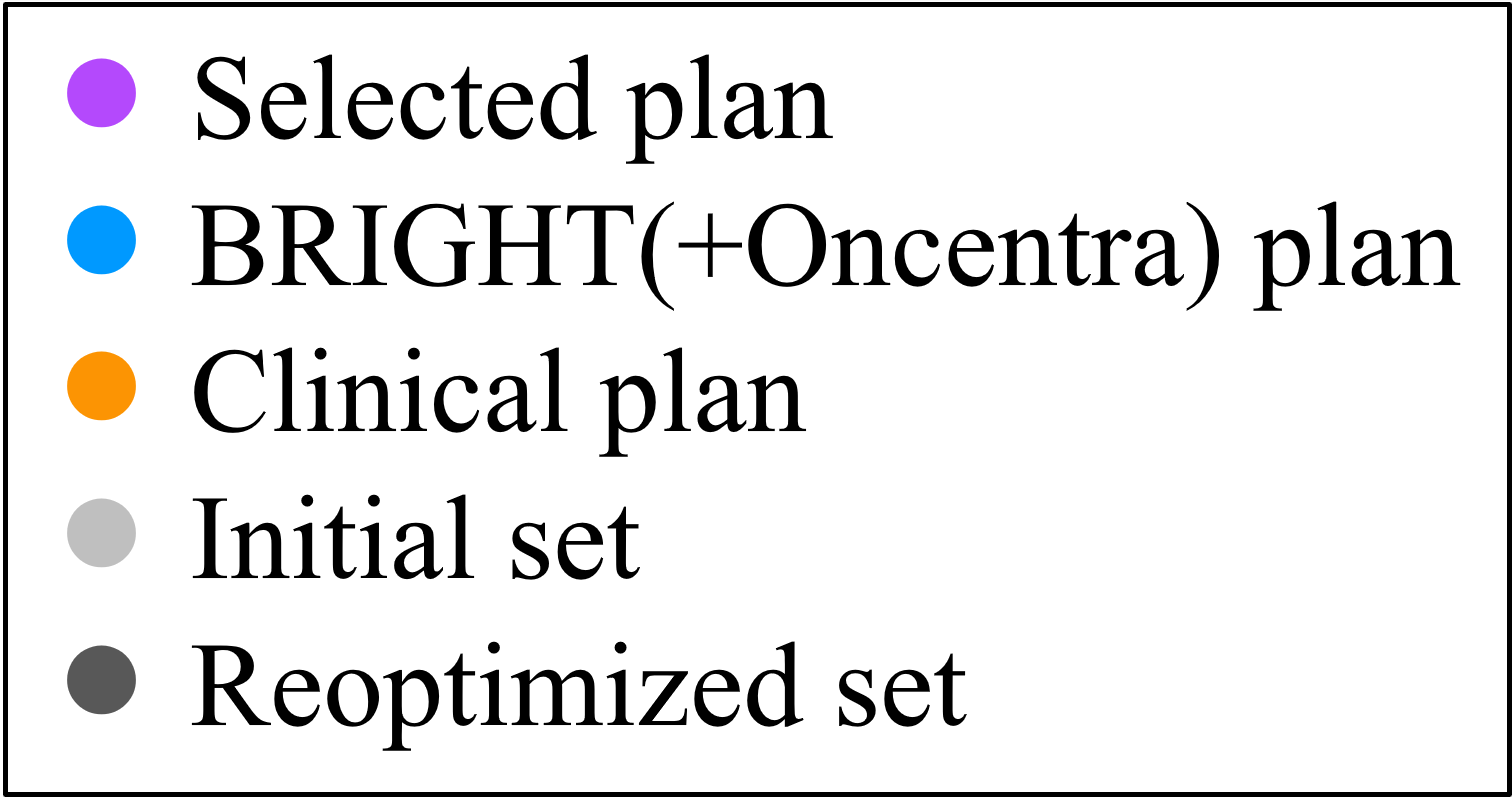}
    \end{minipage}
    \hfill
    \caption{Pareto approximation fronts with the selected plan, possibly adjusted BRIGHT(+Oncentra) plan, and the clinical plan for four representative patient cases. The set of plans that is available after the first optimisation run is plotted in light grey. If re-optimisation was performed and a resulting plan was selected, that set is plotted in dark grey.}
    \label{fig:paretoplots}
\end{figure}

Figure \ref{fig:paretoplots} shows the Pareto approximation front plots with the selected plan (in the BRIGHT GUI), the possibly adjusted final BRIGHT plans, as well as the clinical plan, for four representative patient cases. Results for all other cases can be found in Supplementary Figure 5. Firstly, in 5 out of 10 cases (cases 4, 5, 8, 9, 10), BRIGHT clearly outperforms the clinical plan in terms of LCI and LSI. In the remaining 5 cases, BRIGHT performs at least equally well. 
In 9 out of 10 patient cases, BRIGHT resulted in a treatment plan that met all EMBRACE-II protocol aims (i.e., the plan is located in the Golden Corner). In 4 of these cases, the corresponding clinical plan failed to meet all EMBRACE-II aims.

Manual adjustments in Oncentra were performed in 7 out of 10 cases and generally did not make the plans considerably worse visually, in terms of worst-case DV value in each objective, except in one case (case 10). 
As shown in Table \ref{tab:adjustments}, the median time spent using the BRIGHT GUI was 8.5 min, with a median of 8.0 min spent on adjustments in Oncentra. Per-case statistics are given in Supplementary Table 6 and indicate that the first case required the most time in BRIGHT.

Patient information that was given at the start of the BRIGHT planning sessions had an impact on the plan choice in 4 out of 10 cases. For example, a larger volume of residual tumour could lead the BT team to increase the overall level of coverage when choosing plans (i.e., a higher $\text{CTV}_{\text{HR}}$ dose).
This highlights one advantage of semi-automated treatment planning with multi-plan output: multiple plans with different characteristics are available for selection. This allows the BT team to incorporate patient information (which is not explicitly optimised on) into their final plan choice. 

For 3 out of the 7 cases in which manual adjustments were performed in Oncentra,
the adjustments were thought necessary,
even though all aims from the EMBRACE-II protocol were reached.
According to Questionnaire 1, the reasons for this included underdosage in the first caudal slices, a desire to replace high doses from the needles with the intrauterine applicator, and slightly excessive loading in the intrauterine top.

For other patient cases, adjustments were explored but not necessary (as indicated in Questionnaire 1). Motives for explorations were mostly related to ovoid loadings. In addition, for one of these cases, adjustments that were thought to be needed while the BRIGHT plan was viewed in the BRIGHT GUI, turned out to be superfluous once the same plan was inspected in Oncentra. For another case, the adjustments reversed the improvements made by the BT team during the re-optimisation step in BRIGHT, which focused on minimizing dwell times for certain dwell positions; the adjusted plan was then still used for the rest of the study.

\begin{table}[h]
\footnotesize
    \centering
    \begin{tabular}{l|l}
        Time in BRIGHT & 8.5 min (5.0-16.0) \\
        Time in Oncentra & 8.0 min (0.0-17.0) \\
        Adjustment in Oncentra explored & 7/10 cases \\
        Adjustment considered necessary & 3/10 cases \\
        Patient information impacted plan choice & 4/10 cases
    \end{tabular}
    \caption{Descriptive statistics on plan choice and manual adjustments in Oncentra. Times given as median (minimum-maximum) in minutes.}
    \label{tab:adjustments}
\end{table}

Regarding (a subset of interest of) the features in the BRIGHT GUI, Table \ref{tab:featuresused} shows the number of times they were used. While re-optimisation was performed a median of 1.5 times per patient case, for 2 out of 10 cases the selected BRIGHT plan originated from the initial optimisation run. 
For case 6, the feature was simply not used. For case 5, the team increased the maximum needle contribution as an attempt to reduce the contribution from the intrauterine applicator at a specific location. However, this was not achieved, as the re-optimisation led to higher needle dwell times at different positions than desired by the BT team.
In general, most re-optimisations were related to changing the setting of the maximum dwell time for specific dwell positions and altering the minimum or maximum single and/or total needle contribution. A median of 2 (range: 1-5) plans were marked as favourites per case, which corresponds to median 1 (range: 0-3) per set of generated plans when re-optimisation was used. The difference view for one-on-one plan comparisons was used for half of the cases. Out of these, 3 out of 5 were cases for which no adjustments in Oncentra were explored. The features of ovoid contribution and protocol adjustments not having been utilized could be attributed to the specific patient set which was used in this study.

\begin{table}[h]
\footnotesize
    \centering
    \begin{tabular}{lll}
        Feature & Adjustment type & Number of times used per case \\\hline
        Re-optimisation
           & Maximum time for specific dwell(s) & 1 (0-2) \\
           & Single needle contribution & 0 (0-1) \\
           & Total needle contribution & 0 (0-1) \\
           & Ovoid contribution & 0 (0-0) \\
           & Protocol & 0 (0-0) \\
           & Total & 1.5 (0-3) \\\hline
        \multicolumn{2}{l}{Plans favourited} & 2 (1-5) \\\hline
        \multicolumn{2}{l}{Difference view} & 0.5 (0-2)
    \end{tabular}
    \caption{Median (minimum-maximum) of number of times specific features have been used over all 10 patient cases.}
    \label{tab:featuresused}
\end{table}

\subsection{GUI usability}\label{sec:resGUIusability}

Regarding the usability evaluation according to the SUS \cite{Brooke2020SUS:Scale}, Table \ref{tab:SUSquestionnaire} shows the score per question for all three end-users of the BT team. The mean of the total SUS score can be found in Table \ref{tab:SUSmeans}, and is with 83.3 considered to be 'excellent'. The same table also shows the score per discipline, indicating that the medical physicist and radiation oncologist generally had a higher opinion of the software. All values reflect a score which is considered at least 'good'. The scores per end-user turned out higher than they were for the same end-user in the tuning phase.

The lowest scores were related to the complexity of the system (SUS questions 3 and 7): end-users agreed (rather than strongly agreed) that the system was easy to use and that most users would learn to use the system very quickly. These results highlight the importance of proper training when introducing new software.

Responses to the question whether the end-users fully understood how to use the software prior to the start of the study (software-specific question 1) were either neutral or in agreement. Even though this is one of the worse-scoring questions, end-users also stated that they became considerably more comfortable using the software during the study (software-specific question 2), indicating that with continued use, the software is expected to feel progressively simpler and more intuitive. This is also reflected in question 2 and 9 of the SUS, which states that the system did not feel overly complex, and that end-users felt confident using it. Moreover, there was no inconsistency in the software and the functions were found to be well integrated (SUS questions 5, 6).

It is clear that the end-users furthermore would like to use the system frequently, indicated by the first best-scoring question (SUS question 1). Additionally, they did not find the system cumbersome to use at all (other best-scoring question, SUS question 8).

\begin{table}[h]
\setlength\tabcolsep{3pt}
\footnotesize
    \centering
    \begin{tabular}{llcccc}
        & Statement 
            & \begin{tabular}{@{}c@{}}\ Best\\possible\\score\end{tabular}
            & RO
            & MP
            & RTT\\\hline
        \multirow{10}{*}{SUS} 
            & 1. I think I would like to use this system frequently & 5 & 5 & 4 & 5 \\
            & 2. I found the system unnecessarily complex & 1 & 2 & 1 & 2 \\
            & 3. I thought the system was easy to use & 5 & 4 & 4 & 4 \\
            & 4. I think that I would need the support of a technical person to be able to use this system & 1 & 2 & 1 & 2 \\
            & 5. I found the various functions in this system were well integrated & 5 & 5 & 4 & 4 \\
            & 6. I thought there was too much inconsistency in this system & 1 & 1 & 2 & 2 \\
            & 7. I would imagine that most people would learn to use this system very quickly & 5 & 4 & 4 & 4 \\
            & 8. I found the system very cumbersome to use & 1 & 1 & 1 & 2 \\
            & 9. I felt very confident using the system & 5 & 5 & 4 & 4 \\
            & 10.	I needed to learn a lot of things before I could get going with this system & 1 & 2 & 1 & 2 \\ \hline
        \multirow{4}{*}{\shortstack[l]{Software-\\specific}} & 1. I fully understood how to use the software prior to the start of the study & 5 & 3 & 4 & 3 \\
            & 2. During the study, I became more comfortable using the software & 5 & 5 & 4 & 4 \\
            & 3. I am satisfied with the clinical quality of my selected outcomes & 5 & 5 & 4 & 4 \\
            & 4. I am satisfied with the features available in the software & 5 & 4 & 4 & 4 \\
    \end{tabular}
    \caption{Responses to Questionnaire 2A (SUS) and Questionnaire 2B (software-specific) by the radiation oncologist (RO), medical physicist (MP), and radiation therapy technologist (RTT). Response options range from 1 = strongly disagree, 2 = disagree, 3 = neither agree or disagree, 4 = agree, to 5 = strongly agree. As some statements are worded negatively, the best possible score column indicates whether high or low values are desirable for each statement.}
    \label{tab:SUSquestionnaire}
\end{table}

\begin{table}[h]
\footnotesize
    \centering
    \begin{tabular}{lc}
         & SUS score \\\hline
        Radiation oncologist & 87.5 \\
        Medical physicist & 85.0 \\
        Radiation therapy technologist & 77.5 \\\hline
        Mean & 83.3 \\
    \end{tabular}
    \caption{SUS score (Questionnaire 2A) per discipline, as well as the overall mean.}
    \label{tab:SUSmeans}
\end{table}

In the responses to the open questions, all end-users stated that they find the Golden Corner useful, since it immediately gives an insight into whether all aims are achievable for the patient at hand, or if not, which concessions have to be made. They also all expressed that BRIGHT allows for a much faster start to treatment planning, saving time as compared to manual planning. This was stated to especially be true for difficult patient cases with extensive tumour sizes for which numerous interstitial needles have to be used. Among the features they were satisfied with, all end-users mentioned the re-optimisation, including the adjustable maximum contribution of single dwell positions, specific needles, or all needles. The plan comparison tool (i.e., difference view) was also appreciated by all end-users. Finally, having both the physical dose and the total EQD2 displayed in the GUI was considered helpful, as clinical practice typically requires manual back-and-forth copy-pasting between Oncentra and Excel spreadsheets multiple times during the planning process.

With regard to which features were missing, the end-users solely mentioned the possibility to adjust specific dwell positions manually, without re-optimisation. As improvable, one end-user mentioned that, since the Golden Corner only represents when the aims of the EMBRACE-II protocol have been reached, it might be useful to have a secondary Golden Corner, which reflects the limits of the protocol. This is however not possible since there is no linear relationship between the distance to the aims and the distance to the limits in the worst-case formulation of the objectives as used in BRIGHT. Another comment was that the dwell time chart could also include a contribution per single needle. The most important feedback however was that BRIGHT generates a lot of treatment plans, among which some have small differences. This perception is amplified as a result of the navigation through the set of plans not being reflected by a smooth transition between the isodose lines of subsequent plans, but showing jumping between different dose distributions. This is explained by the fact that plans which have the same DV values, and therefore the same $\mathrm{LCI}$ and $\mathrm{LSI}$ values, can exhibit different 3D dose distributions \cite{Dickhoff2024KeepingBrachytherapy}.

\subsection{Comparison with clinical practice}

\begin{figure}[h]
    \newlength{\h}
    \setlength{\h}{155pt}
    \newlength{\hmin}
    \setlength{\hmin}{150pt}
    \begin{minipage}[t]{\textwidth}
    \centering
    \raisebox{\hmin}[0pt][0pt]{\textbf{a.}}%
    \includegraphics[height=\h]{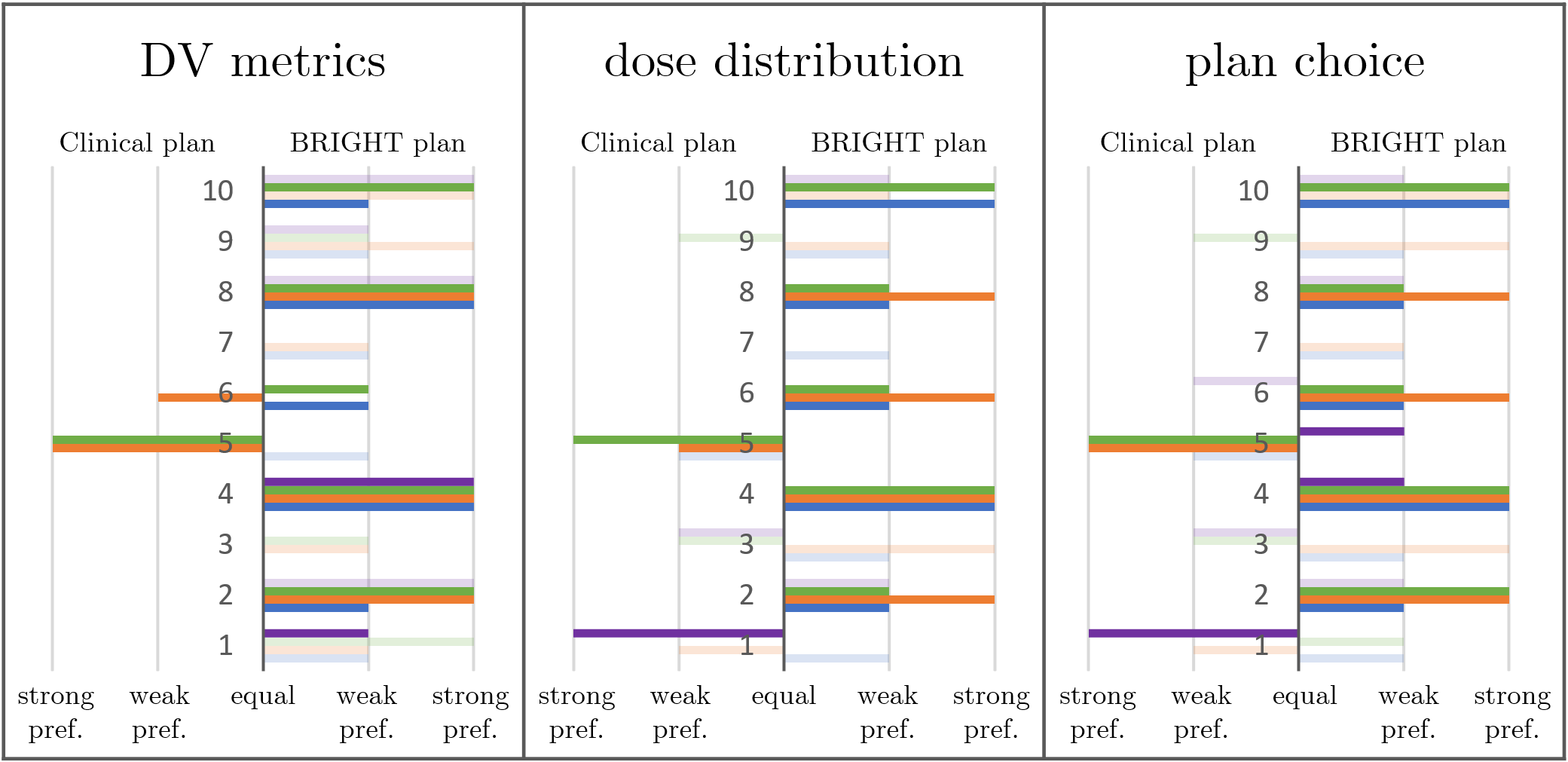}
    \hspace{10pt}
    \raisebox{\hmin}[0pt][0pt]{\textbf{b.}}%
    \includegraphics[height=\h]{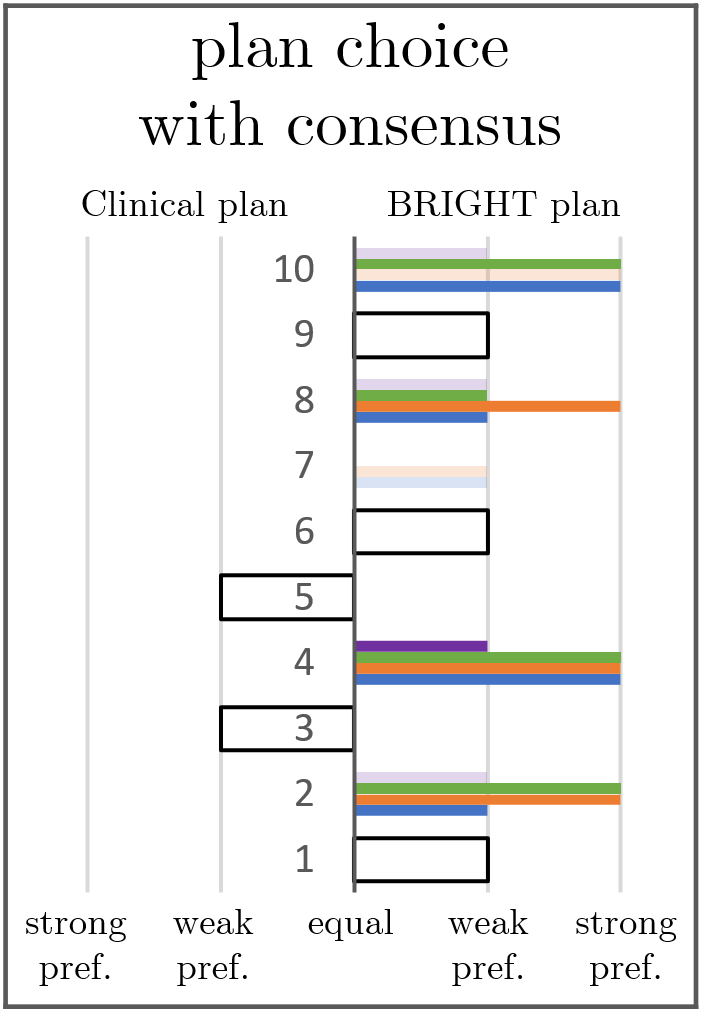}
    \end{minipage}
    \caption{\textbf{a.} Preferred plan (right for BRIGHT, left for clinical) based on EMBRACE-II DV metric values, 3D dose distribution, and overall plan choice for all 10 patient cases. Each colour represents a different end-user, with differences judged as clinically irrelevant shown in transparent. \textbf{b.} Overall plan choice preferences after the consensus meeting, with results plotted as black boxes.}
    \label{fig:boxcharts}
\end{figure}

Results of the blinded comparison of the BRIGHT plan with the clinically used plan for each of the 10 patient cases are presented in Figure \ref{fig:boxcharts}a. One can see that there is prominent interobserver variability, with generally more unanimous preferences for the BRIGHT plan regarding the achieved DV metric values.
For case 5, due to a large $\text{GTV}_\text{res}$ of 31.8 \si{\centi\metre\cubed}, the clinical plan was preferred because of a higher $\text{CTV}_\text{HR} \ \text{D}_{90\%}$ (total EQD2 of 92.7 Gy versus 90.7 Gy in the BRIGHT plan). 
However, a corresponding BRIGHT plan existed that outperformed the clinical plan in all aims, but this plan was not found, as the team did not navigate to higher coverage plans. 
A reason for this could be that there is variability in what is clinically acceptable in a specific moment, as the chosen BRIGHT plan was claimed to be acceptable during the validation session.
Another reason could be that the navigation in the BRIGHT GUI is either improvable - making it more difficult or time-consuming than needed for the users to find the plan that they were looking for in the large set of plans -, or the plan navigation has a learning curve.
The DV metric values and (one slice of) the dose distribution of this plan that was not found during the validation session is presented in Supplementary Table 7 and Supplementary Figure 6. 
For case 1, some end-users criticised that in the BRIGHT plan, a high dose originated from a needle located close to the sigmoid.
This results in a less robust treatment plan, as the sigmoid is known to move, especially when the same plan is used multiple times (in consecutive fractions) based on the same applicator and needle implantation.

Due to the notable interobserver variability, there was a consensus meeting with all four radiation oncologists in which the five cases with diverging overall plan choice preferences (patient cases 1, 3, 5, 6, 9) were discussed. The results are presented in Figure \ref{fig:boxcharts}b. For all of the discussed cases, the radiation oncologists indicated that none of the differences (between the clinical plan and the BRIGHT plan) were considered to be clinically relevant. As such, it can be concluded that the BRIGHT plan was favoured in 8 out of 10 cases, of which 4 with clinically relevant differences. The clinical plan was favoured in 2 out of 10 cases, none with clinically relevant differences.

Regarding the question of whether the plans are, individually, clinically acceptable, the results showed that out of 40 cases (4 end-users $\times$ 10 patient cases), 37 BRIGHT plans (92.5\%) were deemed clinically acceptable, compared to only 30 clinical plans (75\%). Naturally, all of these plans are actually ’clinically acceptable’, since the clinical plans were used for actual treatment, and the BT team also declared that all selected BRIGHT plans were suitable for delivery to the patient. 
Thus, the fact that 25\% of the clinical plans were deemed unacceptable could have stemmed from bias from the comparison, since the question was asked after having selected their preferred treatment plan. Thus, additional insights from having seen what BRIGHT could achieve for the specific patient at hand could have led to a now harsher judgement of the clinical plans. 
Moreover, it may be attributed to interobserver variability, as different end-users have different considerations, and in most cases, the end-users who made the treatment plans were not the same as those who judged the treatment plans.

\vspace{0.5\baselineskip}
\section{Discussion} \label{sec:discussion}

This study presents the first end-user validation of the semi-automated treatment planning method known as BRIGHT, for cervical cancer BT, incorporating full clinical emulation and direct comparison with currently used clinical manual planning. By evaluating BRIGHT in a retrospective setting, yet realistic clinical workflow, our findings provide a critical step beyond previous numerical studies, offering insights into real-world usability. The results demonstrate that semi-automated planning with BRIGHT can achieve treatment quality exceeding, or at least comparable to, current manual approaches,
in clinically feasible planning time.
Importantly, the dedicated GUI developed for BRIGHT was highly appreciated by end-users, as it greatly facilitated navigation through available plan options and enabled intuitive comparison of trade-offs across plans. 
These findings suggest that the use of BRIGHT could enhance treatment plan quality in clinical practice, as well as insights into aim achievability and possible treatment plan trade-offs, supporting its adoption as a viable alternative to manual treatment planning.

The relatively small sample size of ten patient cases could be seen as a limitation of this study. It is worth noting, however, that, firstly, another ten cases were included in the tuning phase of the BRIGHT GUI, as described in the Supplementary Material B. Secondly, although the included patient cases were checked for representativeness, they may be considered more complex, as inclusion was based on having received four fractions of BT. At our department, some patients are treated with three fractions instead of four, as this can improve patient comfort \cite{Williamson2023OutcomesCancer}. The three-fraction scheme is often used when a favourable applicator and needle implantation relative to the target volumes and OARs allows the planning aims to be met within three fractions \cite{Tanderup2020Evidence-BasedCancer, Sturdza2022Image-guidedFractionation}. Therefore, patient cases treated with four fractions are generally more complex, and the cases included in this study have specifically been pointed out as such by the BT team. Thus, feedback on plan quality might have been less extensive if a set of less complex patients would have been included.

Regarding necessary runtimes, even though BRIGHT has now run for 3.7 min to generate the set of plans, this was determined in a worst-case manner \cite{Dickhoff2025MedPhys}, and further speed-ups are still possible. For instance, high-precision dose calculations were used during optimization, though such precision might not be necessary in a clinical setting. In addition, GPU code performance can be further optimized, and even the use of a newer GPU with processor alone could lead to up to a 50\% reduction.

Another point worth discussing is the importance of the quality of the delineations for (semi-)automatic treatment planning. When evaluating a treatment plan, radiation oncologists can disagree with the given delineations and, instead of adjusting the delineations, they do or do not take into consideration the specific subregions in the evaluation of the treatment plan. Furthermore, they could intuitively consider subregions within the delineations differently due to the underlying anatomical structures. For the optimisation algorithm in BRIGHT, however, the information provided is interpreted as is, and - unless explicitly specified as an ROI - it does not differentiate between more or less important parts, or regions with more or less certainty. Small variations in delineations can have a large impact on dosimetric parameters \cite{Duane2014ImpactBrachytherapy, Damato2014DosimetricBrachytherapy}, and these effects can be amplified in the automatic optimisation of a treatment plan.

Furthermore, a factor that could have impacted the results of this validation study in favour of the clinical plans is that it might be the case that more care was taken in optimising the clinical plans as compared to the BRIGHT plans. 
It stands to reason that when patient care is at stake, more effort and vigilance is granted to the details of the treatment plan. This is in accordance with other studies, which realized that treating physicians are significantly less critical in retrospective settings as compared to a real-world clinical setting \cite{McIntosh2021ClinicalCancer}. We aimed to minimize this effect by restricting the BRIGHT validation sessions to a maximum of one hour, ideally shorter, to accommodate the participants' busy schedules, and most of all, by emulating the exact steps and information available during clinical treatment planning.

Moreover, the BT team did not hesitate to adjust the treatment plans in Oncentra, since, in clinical practice, the treatment plans would have to pass through Oncentra anyway as the current radiation machine is vendor specific and can only be used through its associated regulatory-approved treatment planning system. Thus, knowing that there is a subsequent check as well as room for possible adjustments in Oncentra could have impacted the time the BT team was willing to spend in BRIGHT.

Future research should focus on improving the navigation and selection of treatment plans within the GUI to enhance usability. A more intuitive transition between plans that are neighbouring in objective space (i.e., along the approximation front) - particularly in isodose lines - could make navigation smoother and plan selection more efficient. 
Furthermore, the dose calculation engine can be extended to include model-based dose calculation \cite{model-based-dc}, offering enhanced accuracy and personalization.
The supported imaging modalities shown in the GUI could be broadened to incorporate Positron Emission Tomography.
Additionally, while this study focused on cervical cancer BT, a first version of this software is used in clinical practice for prostate BT, and it can also be applied to other applications such as breast BT \cite{Bouter2025AI-basedBRIGHT}, though its use in this case still needs to be evaluated.

This study represents the first interaction of end-users with BRIGHT enhanced with a custom-developed GUI for cervical cancer BT. It thereby provides extensive insight into considerations and necessities linked to moving from manual to semi-automatic treatment planning. The latter involves a different way of thinking than manual planning does, as the users’ work now switches from building treatment plans step by step to navigating a set of plans in order to find the preferable plan for the patient at hand. Discussions with some BT professionals of our department during this study have revealed that this might lessen the enjoyment they get from their work and semi-automatic treatment planning can therefore generally be met with scepticism and criticism.
There is still an important step to be taken here, and this should not be underestimated in the broader implementation of automated treatment planning.
This might be the reason that radiation therapy technologists generally seem to score the software lower than the radiation oncologists and the medical physicists (see Section \ref{sec:resGUIusability} and Supplementary Table 2), as they are the ones in our department that would otherwise make the treatment plan manually. Another reason can generally be natural variations in people's critical disposition.

Increasing the users’ trust in the system can be facilitated by multi-objective optimisation in itself, as it allows the users to gain insights into what is achievable for each patient by observing numerous approximately Pareto-optimal plans that each represent a different near-best coverage-sparing trade-off. It therefore differs vastly from deep learning or machine learning approaches, which present just one solution without giving insights into concessions made. General consensus with regard to Artificial Intelligence implementations is that it should currently always be checked by an expert \cite{Callens2024IsOff}, which is intrinsic to BRIGHT as the final plan choice is left to the expert.

One of the main points of feedback in this study was the desired inclusion of a feature that allows for manual adjustments of the dwell times, i.e., modifications without subsequent re-optimisation (as was done during this study in Oncentra). However, this would inevitably result in mathematically sub-optimal treatment plans, as any manual adjustment causes the plan to deviate from the Pareto approximation front. 
There is no proof that these manual adjustments would lead to a better clinical outcome.
The only published clinical studies are within the EMBRACE-II framework and relate maximum and minimum DV metric values to OAR toxicity levels, side effects, and tumour control \cite{EMBRACE}. BRIGHT does obtain the best possible treatment plans in terms of these DV metrics since it directly optimises on them.

However, allowing for adjustments to the automatically generated treatment plans (both using the BRIGHT GUI and Oncentra) at this stage can act as a key quality assurance tool in order to build trust in the automatically generated plans by allowing the users to explore plan modifications that they would habitually attempt. 
Some ’choices’ made by BRIGHT were initially criticised but were later accepted after exploring alternative manual adjustments, which failed to achieve the desired outcomes (see Section \ref{sec:results:planning}). The BT team then indicated that they needed some time before agreeing with BRIGHT. This suggests that users may not immediately trust automatic treatment planning upon first use, particularly in patient cases with irregularities, such as unusual tumour location or size, or suboptimal applicator implantation  (e.g., ovoids placed too low to properly align with the cervix).

We decided to focus on the main strength of semi-automatic treatment planning, which is that it quickly approximates the mathematically best possible plan(s), while still allowing re-optimisation to tailor the treatment plans to the specific patient and user. After this step, further local adjustments can be done manually outside of BRIGHT if needed. Enabling re-optimisation at this stage is essential, given the growing belief that there might not be one clinical protocol for automatic treatment planning that is satisfactory for the vast majority of patients, especially between different institutes \cite{Fiandra2024Multi-centreRadiotherapy}. 
A similar point can be made from the interobserver variation in treatment plan evaluation as seen in this study. This is a known issue in radiotherapy, due to, among others, subjective considerations of the importance of some planning aims, growing experience over time, and/or diverging semantic understanding (e.g., of 'clinical acceptance') \cite{Vanderstraeten2018AutomatedPreference}.
While some studies suggest to develop separate automatic planning models according to each expert's preferences \cite{Wang2024ClinicalCancer}, we believe that semi-automatic treatment planning can aid in 
unifying treatment planning protocols, while still allowing the experts to
find the high-quality treatment plan that fits most with their preferences, for each individual patient.
Having the final plan choice in multi-objective plan navigation, as well as the possibility to apply small manual adjustments, gives the expert room to include their own expertise, and - much more importantly - tailor the planning to patient-specific needs.

Finally, the aim of this study was to explore the potential of BRIGHT for clinical use.
While plan preference outcomes may vary on a case-by-case basis, these variations are inherent to clinical practice, such as the interobserver variability in terms of plan choice as well as desired plan adjustments as seen in the comparisons in this study. All plans were classified as being clinically acceptable at the time of creation, and any further critique was influenced by individual clinical perspectives and variability. 

\vspace{0.5\baselineskip}
\section{Conclusion}

BRIGHT was enhanced with a custom-developed GUI for navigation through a Pareto approximation set of multiple brachytherapy treatment plans, each representing a different trade-off between target volume coverage and OARs sparing. We conducted for the first time an end-user validation study of BRIGHT for cervical cancer BT on ten patient cases representative of current clinical practice. The GUI was found to be an ’excellent’ system overall, according to the SUS system score. BRIGHT was appreciated as a semi-automatic treatment planning method, with a Golden Corner that indicates at one glance whether the patient at hand is a straightforward case in which all planning aims can be achieved, or whether concessions must be made. 
The ability to re-optimize treatment plans within BRIGHT as well as the option to compare plans pairwise, were highly valued. However, plan navigation could be improved to make it more intuitive by enabling the isodose lines of neighbouring plans along the approximation front to transition smoothly. \par
Direct comparison of the BRIGHT and clinical plans showed some variability between end-users, which we consider to be inherent to clinical practice. The comparison also revealed that the BRIGHT plan was preferred for most patient cases, with some differences being deemed clinically relevant.

\section*{Disclosure}

This work was supported by the Dutch Cancer Society [KWF Kankerbestrijding; project number 12183] and by Elekta Brachytherapy, Veenendaal, The Netherlands.

\section*{Acknowledgements}
The brachytherapy team of the Department of Radiation Oncology at the Leiden University Medical Center is gratefully acknowledged for their support and involvement in this study.\\


\printcredits

\bibliographystyle{unsrturl_mod}

\bibliography{references-mod}

\bio{}
\endbio

\end{document}